%% file: emnlp2020.tex
\documentclass[11pt,a4paper]{article}
\usepackage[hyperref]{emnlp2020}
\usepackage{times}
\usepackage{latexsym}

\usepackage{microtype}

\usepackage{times}

\usepackage{soul}
\usepackage{url}
\usepackage{graphicx}
\usepackage{amsmath}
\usepackage{booktabs}
\usepackage{footnote}
\usepackage{array}
\usepackage{soul}
\usepackage{color}
\usepackage[utf8]{inputenc}
\usepackage{amsmath}
\usepackage{multirow}
\usepackage[linesnumbered,boxed,ruled,commentsnumbered]{algorithm2e}
\newcommand{\figref}[1]{Figure \ref{#1}}
\newcommand{\eqnref}[1]{Eq. \ref{#1}}
\newcommand{\tabref}[1]{Table \ref{#1}}
\newcommand{\secref}[1]{Section \ref{#1}}
\newcommand{\algoref}[1]{Algorithm \ref{#1}}

\aclfinalcopy 
\def\aclpaperid{215} 

\title{Multi-turn Response Selection using Dialogue Dependency Relations}

\author{Qi Jia$^1$ \hspace*{1cm}
	Yizhu Liu$^2$ \hspace*{1cm}
  	Siyu Ren$^3$ \hspace*{1cm} Kenny Q. Zhu$^4$\thanks{\hspace{2mm}The corresponding author.}\\
  Shanghai Jiao Tong University \\
  Shanghai, China \\
  \texttt{\{$^1$Jia\_qi,$^2$liuyizhu,$^3$roy0702\}@sjtu.edu.cn} \hfill
  \texttt{$^4$kzhu@cs.sjtu.edu.cn} \\ 
  \AND
  Haifeng Tang \\
 China Merchants Bank Credit Card Center\\
  Shanghai, China \\
  \texttt{thfeng@cmbchina.com} \\
}

\date{}	

\begin{document}
\maketitle
\begin{abstract}
Multi-turn response selection is a task designed for developing dialogue agents. The performance on this task has a remarkable improvement with
pre-trained language models. However, these models simply concatenate the turns in dialogue history as the input and largely ignore the dependencies between the turns. In this paper,  we propose a dialogue extraction algorithm to transform a dialogue history into threads based on their dependency relations. Each thread can be regarded as a self-contained sub-dialogue. We also propose Thread-Encoder model to encode threads and candidates into compact
representations by pre-trained Transformers and finally get the matching score through an attention layer. The experiments show that dependency relations are helpful for dialogue context understanding, and our model outperforms the state-of-the-art baselines on both DSTC7 and DSTC8*, with competitive results on UbuntuV2.
\end{abstract}
\input{intro}

\input{method}

\input{eval}

\input{results}

\input{relatedwork}

\input{conclusion}

\section*{Acknowledgement}
This research was supported by the SJTU-CMBCC Joint Research Scheme, 
SJTU Medicine-Engineering Cross-disciplinary Research Scheme, and NSFC
grant 91646205.

\bibliographystyle{acl_natbib}
\bibliography{emnlp2020}

\end{document}

%% file: intro.tex
\section{Introduction}

Dialogue system is an important interface between machine and human.  An intelligent dialogue agent is not only required to give the appropriate response based on the current utterance from the user, but also consider the dialogue history. Dialogue context modeling has been a key point for developing such dialogue systems, including researches on state tracking~\cite{abs-1907-01669,RenNM19}, topic segmentation~\cite{NanDNX19,kim2019dynamic}, multi-turn response selection~\cite{TaoWXHZY19,GuLL19}, next utterance generation~\cite{abs-1911-00536,ChenCQYW19}, etc. In this paper, we target on the multi-turn response selection task, which is first proposed by Lowe et al.~\shortcite{LowePSP15} and is also a track in both DSTC7~\cite{gunasekara2019dstc7} and DSTC8~\cite{dstc8}.

Given a dialogue history made up of more than one utterance, the selection task is to choose the most possible next utterance from a set of candidate responses.  Previous work on this task can be roughly divided into two categories: sequential models and hierarchical models. The former ones, including \cite{LowePSP15,YanSW16,abs-1901-02609}, concatenate the history utterances into a long sequence, try to capture the similarities between this sequence and the response and give a matching score. The latter ones, including \cite{TaoWXHZY19,WangWC19,GuLL19}, extract similarities between each history utterance and the response first. Then, the matching information is aggregated from each pair 
(mostly in a chronological way) to get a final score. 
There is little difference between the performance of these two kinds of 
architectures until the emergence of large pre-trained language models.
 
 \begin{figure}
 	\centering
 	\includegraphics[scale=0.39]{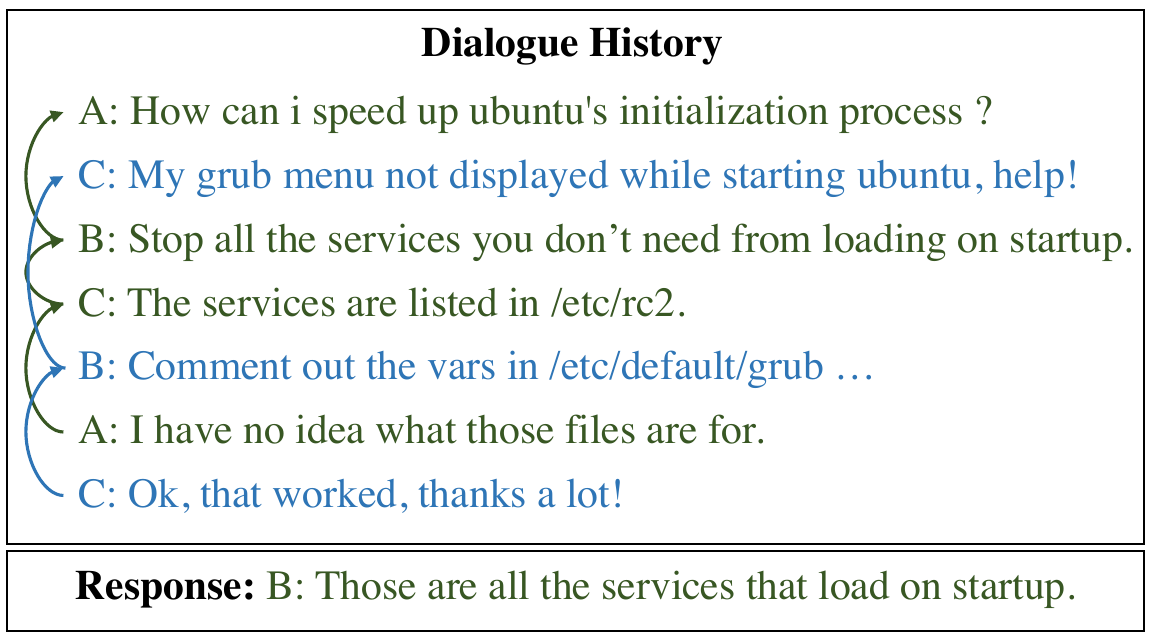}
 	\caption{An example of the tangled dialogue history. A, B and C are three participants. Texts in different colors represent different dialogue threads.}
 	\label{fig:example}
 \end{figure}

Work such as \cite{abs-1908-04812,vig2019comparison} has shown the 
extraordinary performance of the pre-trained language models on dialogues. 
These pre-trained models are easily transferred to the response selection 
task by concatenating all of the utterances as the input. 
All of the words in dialogue history can directly interact with each other 
via transformers like Bi-encoder, even the words both in the dialogue history 
and the candidate response if time permits, such as Cross-encoder~\cite{humeau2019poly}. 
However, since such models can be regarded as the ultimate 
architecture of the sequential-based models, the dialogue dependency 
information between the utterances is largely ignored due to the 
concatenation operation~\cite{WuWXZL17}. An example is shown in Figure \ref{fig:example}. The dependency relations can definitely help us to understand the two tangled dialogue threads.
Besides, we always need to truncate the earlier dialogue history 
to limit the size of the model and make the computation efficient. 
However, it isn't always that the nearest utterances are more important. 
As we can see in Figure \ref{fig:example}, several dialogue threads may 
be tangled especially in multi-party chat rooms, 
it's hard to tell which dialogue thread will be moving on.

In this paper, we propose to incorporate dialogue dependency information 
into the response selection task. 
We train a dialogue dependency parser
to find the most probable parent utterance for each utterance in a session. We name such relation between utterances as ``reply-to''.
Then, we empirically design an algorithm to extract dialogue threads, which is represented by a path of dependency relations according to the parsed trees.
The extracted threads are sorted by the distance between the final utterance in each 
thread and the response in ascending order, following the intuition that the closer utterances are more relevant.
After that, we propose the model named Thread-Encoder based on a pre-trained language 
model.
Each encoder in the model can distill the critical information from each dialogue thread or the candidate response. 
Finally, another attention layer is used to calculate the matching score 
with thread representations and the candidate representation. The candidate with the highest matching score will be selected as the final response.

We collect the training data for dialogue dependency parser from a dialogue 
disentanglement dataset~\cite{KummerfeldGPAGG19} in the technical domain. 
And we do response selection experiments among UbuntuV2, DSTC7 and DSTC8*. 
These datasets consist of dialogues in the same domain but under different settings, 
including two-party dialogues and multi-party dialogues. 
The results demonstrate our model's strong capability to represent multi-turn
dialogues on all of these datasets.

Our main contributions are as follows:
\begin{itemize}
	\item As far as we know, we are the first to incorporate dialogue 
dependency information into response selection task, 
demonstrating that the dependency relations in the dialogue history 
are useful in predicting dialogue responses (Sec \ref{sec:relatedwork}). 
	\item Based on the predicted dependencies, we design a straight-forward but effective algorithm to extract several threads from the dialogue history  (Sec \ref{sec:DSA}). 
The results show the algorithm is better than other simple segmentation 
methods on the response selection task.
	\item We propose the Thread-Encoder model, incorporating dialogue 
dependency information by threads and utilizing the pre-trained language 
model to generate corresponding representations (Sec \ref{sec:tem}). The experimental results 
show that our model outperforms the state-of-the-art baselines on 
DSTC7 and DSTC8* datasets, and is very competitive on UbuntuV2 (Sec \ref{sec:ra}).
\end{itemize}

%% file: method.tex
\section{Approach}

The multi-turn response selection tasks represent each dialogue as a triple $T=\langle C, R, L\rangle$, where $C=\{t_1, t_2,...,t_n\}$ represents the history turns. $R$ is a candidate response and $L$ is the $0/1$ label indicating whether $R$ is the correct response or a negative candidate. To incorporate the dependency information between the history turns, we design a straight-forward algorithm to extract the dialogue history $C$ into dialogues threads $\langle C_1, C_2, ..., C_M\rangle$ based on the predicted dependencies, along with an elaborately designed model to find the function $f(C_1, C_2, ..., C_M, R)$, which measures the matching score of each $(C, R)$ pair. Both the extraction algorithm and the model will be explained as follows.

\subsection{Dialogue Extraction Algorithm}
\label{sec:DSA}
Since it's impossible for the large pre-trained language models to take all of the dialogue history turns as the input under the computational power nowadays, these models usually set a truncate window and only consider the top-k most recent turns or tokens. However, several dialogue threads may exist concurrently in two-party~\cite{DuPX17} or multi-party dialogues~\cite{TanWGWPGCY19}. Such coarse-grained truncating operation may not only bring in redundant turns from 
other dialogue threads, but also exclude the expected turns given earlier 
in the current dialogue thread, hurting the representation capability of pre-trained language models. Extracting the whole history into self-contained dialogue threads can help preserve more relevant turns and avoid the negative effects of encoding irrelevant turns by a single language model.

Motivated by the above, we aim to analyze the discourse structures in dialogue history at first. We utilize the discourse dependency parsing model for dialogues proposed by Shi and Huang~\shortcite{ShiH19}. It is a deep sequential model that achieves the state-of-the-art performance on the STAC corpus.
Instead of predicting the predefined relation types between Elementary Discourse Units(EDUs), we borrow the proposed model in this work to find if there exist dependency relations between utterances in the given dialogue history.
The model scans through the dialogue history and predicts the most likely parent turn for each turn. 
It finally constructs a dependency tree for each dialogue history with a confidence score 
on each edge.

\begin{algorithm}
	\scriptsize
	\SetAlgoNoLine 
	\SetKwInOut{Input}{\textbf{Input}}
	\SetKwInOut{Output}{\textbf{Output}} 
	
	\Input
	{
		The dependency tree $T$ with confidence scores on each edge $e_{ji}=(t_i, t_j, P_{ji})$, where $i.j=1, 2, ..., n$ and $j>i$;\\
		The threshold for the confidence score $P$.
	}
	\Output{
		The threads $C'=\langle C_1, C_2, ..., C_M\rangle$, and each is made up of a sequence of turns.
	}
	\BlankLine
	
	\For {$e_{ji}$ in $T$}{
		\If{$P_{ji}<P$}{delete $e_{ji}$ from $T$}
	}
	The forest $T' = T$\\
	The set of threads $C'=\emptyset$\\
	\For{each leaf node in $T'$ }{
		$C_{tmp} =$ all the node from the leaf node to the corresponding root.\\
		$C'=C'\cup C_{tmp}$
	}
	Rank the threads in $C'$ based on the index of the leaf node in descending order.
	
	\caption{The Dialogue Extraction Algorithm\label{alg:A}}
	\label{alg:DSA}
\end{algorithm}

\begin{figure}
	\centering
	\includegraphics[scale=0.40]{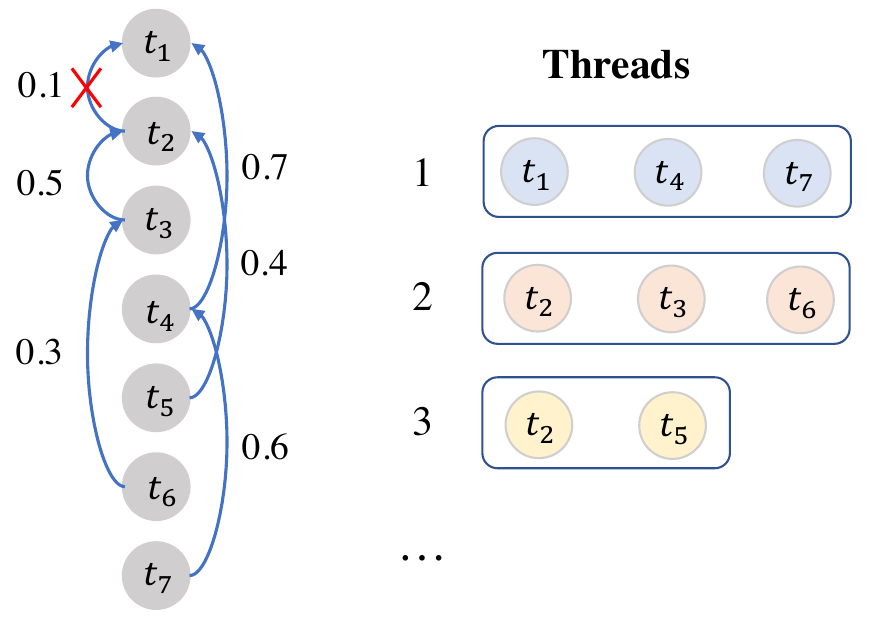}
	\caption{An example of the algorithm when the threshold $P=0.2$. The figures are the confidence scores of corresponding predicted dependency relations.}
	\label{fig:algorithm}
\end{figure}

Then, the dialogue extraction algorithm is designed to extract original long history into dialogue threads according to dependency tree $T$ and confidence scores. The algorithm is 
depicted in \algoref{alg:A}. $e_{ji}$ is a directed edge with head $t_i$ and tail $t_j$, 
indicating that turn $j$ is a reply of turn $i$ with probability $P_{ji}$.
The threshold $P$ is a hyper-parameter. It is noteworthy that we still follow the intuition 
that the turns closer to the responses are more likely to be useful than others. 
As a result, the threads are returned in ascending order according to the distance between 
the last turns in each thread and the response. 
An illustration of the algorithm is shown in Figure \ref{fig:algorithm}. The 7-turn dialogue history is extracted into three threads.

\subsection{Thread-based Encoder Model}
\label{sec:tem}
In the work from Humeau et al.~\shortcite{humeau2019poly}, they use a pre-trained language model as the context encoder and generate the embedding for dialogue history. Inspired by this work, we also utilize pre-trained language models to encode natural texts into meaningful representations.

Given the extracted self-contained dialogue threads $\langle C_1, C_2, ..., C_M \rangle$, we utilize a pre-trained language model to encode the content of each dialogue thread in parallel and another pre-trained language model to encode the candidate respectively. 
If the candidate representation matches well with one or more thread representations, 
that candidate is probably the correct response. 

\begin{figure}
	\centering
	\includegraphics[scale=0.34]{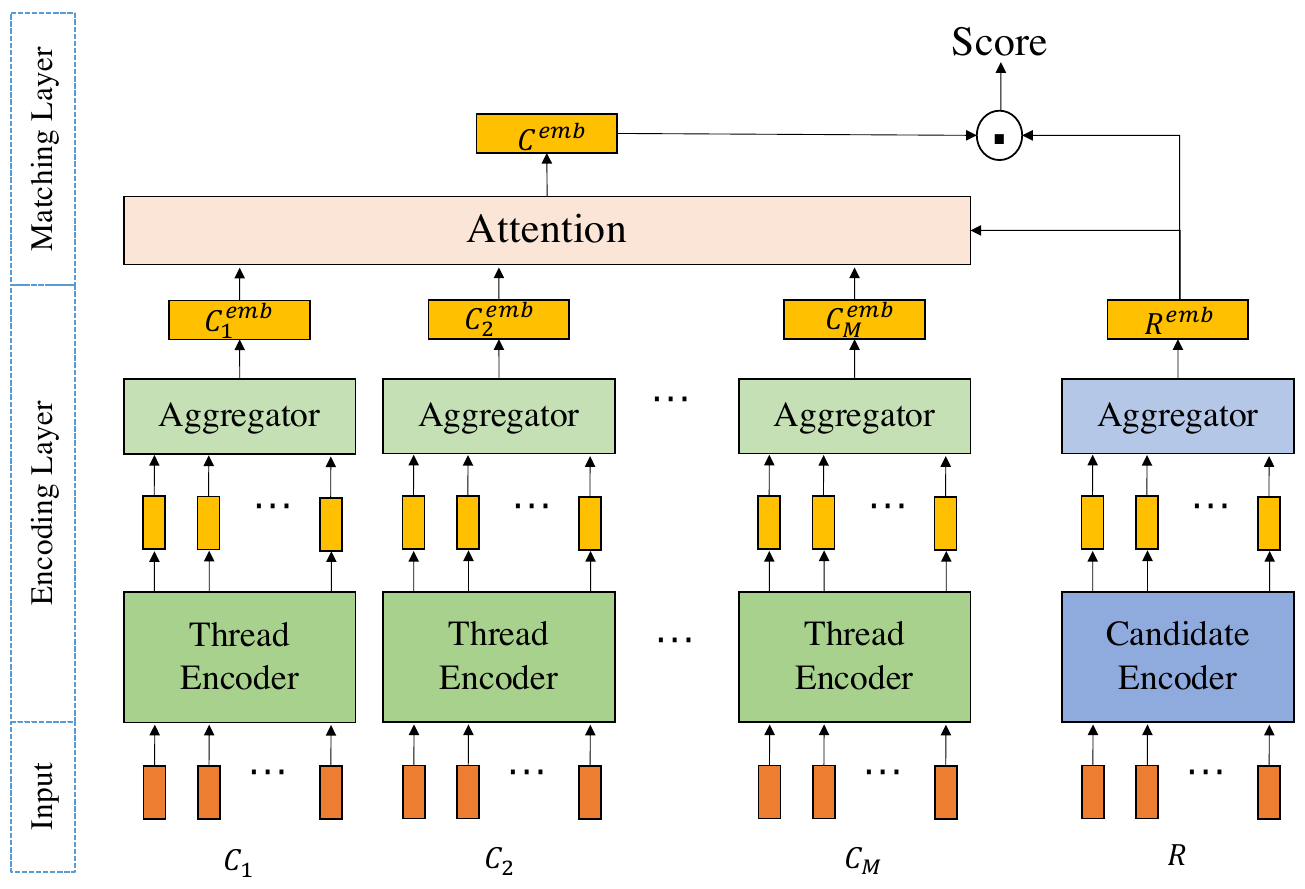}
	\caption{The architecture of Thread-Encoder model. All of the blocks in the same color share parameters.}
	\label{fig:model1}
\end{figure}

The architecture of our model \textbf{Thread-Encoder} (shown in \figref{fig:model1}) can be divided into two layers: Encoding Layer and Matching Layer.

\subsubsection{Encoding Layer}
We use the pre-trained language model released by Humeau et al.~\shortcite{humeau2019poly}. This large pre-trained Transformer model has the same architecture as BERT-base \cite{DevlinCLT19}. It has 12 layers, 12 attention heads and 768 hidden size. Different from the original one trained on BooksCorpus and Wikipedia, the new language model is further trained on Reddit~\cite{abs-1904-06472}, a large dialogue dataset with around 727M context-response pairs. The pretraining tasks include masked language model and next utterance prediction~\footnote{``Utterance'' and ``turn'' are 
interchangeable in this paper.}. Finally, the pre-trained model can be used for 
a wide range of multi-sentence selection tasks with fine-tuning.

In our model, the encoder layer uses two Transformers, thread encoder $T_1(\cdot)$ and candidate encoder $T_2(\cdot)$, both initialized with the pre-trained weights. $T_1(\cdot)$ is used for encoding threads, and all of the turns in a thread are concatenated into a long sequence in chronological order as the input. 
$T_2(\cdot)$ is used for encoding the candidate. The inputs to the 
Transformer encoder are surrounded by the special token $[S]$, 
consistent with the operations during pretraining.

Above the Transformer encoder is an aggregator $agr(\cdot)$ that aggregates
a sequence of vectors produced by the encoder into one or more vectors. 
In a word, the threads and response can be encoded as follows:

\begin{equation}
\begin{aligned}
C^{emb}_m&=agr_1(T_1(C_m))\\
R^{emb}&=agr_2(T_2(R)),
\end{aligned}
\label{eq:agr}
\end{equation}
where $m=1,2,...M$ and $M$ is the number of dialogue threads. For $arg_1(\cdot)$, if we simply use "average" function for the aggregator, only one representation will be encoded for each thread. We name this model as \textbf{Thread-bi}. If we use "multi-head attention" as the aggregator, multiple representations will be encoded for each thread. We name this model as \textbf{Thread-poly}. The aggregator $agr_2(\cdot)$ for candidate representations is the average over input vectors.

\subsubsection{Matching Layer}

Given the encoded threads $\langle C^{emb}_1, C^{emb}_2, ..., C^{emb}_M\rangle$ and candidate $R^{emb}$, we further use an attention layer to distill the information from the threads by attending the query $R^{emb}$ to each $C_m^{emb}$:

\begin{equation}
\begin{aligned}
C^{emb}&=\sum_{m=1}^Mw_mC^{emb}_m
\end{aligned}
\end{equation}
where

\begin{equation}
\begin{aligned}
s_m &= (R^{emb})^\top \cdot C^{emb}_m\\
w_m &= \exp(s_m)/\sum_{k=1}^M\exp(s_k)
\end{aligned}
\end{equation}

The final matching score is given by:
\begin{equation}
\begin{aligned}
S = F(C_1, C_2, ..., C_M, R) = (R^{emb})^\top\cdot C^{emb}
\end{aligned}
\label{eq:matching_score}
\end{equation}

We consider the other correct responses in a mini-batch as the negative candidates to accelerate the training process~\cite{MazareHRB18}. The whole model is trained to minimize the cross-entropy loss as follows:
\begin{equation}
\begin{aligned}
loss = - \frac{1}{A}\sum_{a=1}^{A}\sum_{b=1}^{A} L_{ab}\log(\frac{\exp(S_{ab})}{\sum_{c=1}^{A}\exp(S_{ac})})
\end{aligned}
\end{equation}
where $A$ is the batch size. $L_{ab}$ equals $1$ when $a=b$, otherwise $0$. $S_{ab}$ is a matching score based on \eqnref{eq:matching_score}.

%% file: eval.tex
\section{Experimental Setup}
\label{sec:eval}
In this section, we introduce the datasets, baselines and implementation details of our model\footnote{The codes and data resources
can be found in https://github.com/JiaQiSJTU/ResponseSelection.}.

\subsection{Datasets}
Our experiments are performed on three datasets:
UbuntuV2, DSTC 7 and DSTC 8*.
\itemsep0em
\begin{itemize}

\item \textbf{UbuntuV2}~\cite{LowePSCLP17} consists of two-party dialogues extracted from the Ubuntu chat logs.

\item \textbf{DSCT7}~\cite{gunasekara2019dstc7} refers to the dataset for DSTC7 subtask1 consisting of two-party dialogues.

\item \textbf{DSCT8*} refers to the dataset for DSTC8 subtask 2, containing dialogues between multiple parties. We remove the samples without correct responses in the given candidate sets~\footnote{We do this to eliminate the controversy of solving no correct response in different ways and try to focus on dialogue context modeling.}.
\end{itemize}
More details of these three datasets are in Table \ref{tab:dataset}.

\begin{table}[th]
	\centering
	\small
	\begin{tabular}{lrrr}
		\toprule[1pt]
		\textbf{} & {UbuntuV2}& {DSTC7}& {DSTC8*} \\ 
    	\midrule[1pt]
		{Train} & 957,101 &100,000&89,813\\
		{Valid} & 19,560 & 5,000&7,660\\
		{Test} &18,920 &1,000&7,174\\
		{\#Candidates} &10& 10&100\\
		{\#Correct Response}&1&1&1 \\
		{\#Turns } & 3-19&3-75& 1-99\\
		\bottomrule[1pt]
	\end{tabular}
	\caption{The statistics of the datasets used in this paper.}
	\label{tab:dataset}
\end{table}

\subsection{Baselines}
We introduce several state-of-the-art baselines to compare with our results as follows.
\begin{itemize}
	 \item \textbf{DAM}~\cite{WuLCZDYZL18} is a hierarchical model based entirely on self and cross attention mechanisms.
	\item \textbf{ESIM-18}~\cite{abs-1802-02614} and \textbf{ESIM-19}~\cite{abs-1901-02609} are two sequential models, which are the modifications and extensions of the original ESIM~\cite{ChenZLWJI17} developed for natural language inference. The latter one ranked top on DSTC7.
	\item \textbf{IMN}~\cite{GuLL19} is a hybrid model with sequential characteristics at matching layer and hierarchical characteristics at aggregation layer.
	\item \textbf{Bi-Encoder} (Bi-Enc), \textbf{Poly-Encoder} (Poly-Enc) and \textbf{Cross-Encoder} (Cross-Enc)~\cite{humeau2019poly} are the state-of-the-art models based on pre-trained model. 
\end{itemize}

\subsection{Implementation Details }
According to Section \ref{sec:DSA}, we firstly transform the dialogue disentanglement dataset~\cite{KummerfeldGPAGG19}.
Turns are clustered if there exists a ``reply-to'' edge, 
and we obtain 4,444 training dialogues from the original training set and 480 test dialogues 
from the original valid set and test set. Only 7.3\% of turns have multiple parents. 
Since the parsing model can only deal with dependency structure with a single parent, 
we reserve the dependency relation with the nearest parent in these cases. 
We trained a new parser on this new dataset. The results on the new test set are shown 
in \tabref{tab:parser}. It shows that in-domain data are useful for enhancing the 
results for dialogue dependency prediction.

\begin{table}[th]
	\centering
	\small
	\begin{tabular}{lccc}
		\toprule[1pt]
		\textbf{} & {Precision}& {Recall}& {F1} \\ 
		\midrule[1pt]
		{Trained on STAC} & 67.37 &64.43&65.86\\
		{Trained on the new dataset} & 71.44 & 68.32&69.85\\
		\bottomrule[1pt]
	\end{tabular}
	\caption{The results of the dialogue dependency parser.}
	\label{tab:parser}
\end{table}

For the response selection task, we implemented our experiments based on ParlAI~\footnote{\url{https://github.com/facebookresearch/ParlAI}}. Our model is trained with Adamax optimizer. The initial learning rate and learning rate decay are $5\mathrm{e}{-5}$ and $0.4$ respectively. The candidate responses are truncated at $72$ tokens, covering more than $99\%$ of them. The last $360$ tokens in the concatenated sequence of each thread are reserved. The BPE tokenizer was used. 
We set the batch size as $32$. The model is evaluated on valid set every $0.5$ epoch. 
The training process terminates when the learning rate is $0$ or the hits@1 on validation no longer increases within $1.5$ epochs. 
The threshold in the Algorithm~\ref{alg:DSA} is set to $0.2$ and we preserve at most top-$4$ threads for each sample, avoiding the meaningless single turns while ensuring the coverage of original dialogue contexts. The results are averaged over three runs.
For UbuntuV2 and DSTC7 training set, we do data augmentation: each utterance of a sample can be regarded as a potential response and the utterances in the front can be regarded as the corresponding dialogue context.

Our experiments were carried out on 1 to 4 Nvidia Telsa V100 32G GPU cards. The evaluation metrics for response selection are hits@k and MRR, which are widely used and the codes can be found in ParlAI.

%% file: results.tex
\section{Results and Analysis}
\label{sec:ra}
Here we show results on dialogue thread extraction and response selection of the three datasets, and give some discussions on our model design.
\subsection{Extraction Results}
\label{sec:extrresults}
\begin{table*}[th!]
	\centering
	\small
	\begin{tabular}{lc|ccc|cccc}
		\toprule[1pt]
		\multicolumn{2}{c}{Dataset} &  avg\#thd & avg\#turn & std\#turn & 1-thd(\%) & 2-thd(\%)& 3-thd(\%) & 4-thd(\%)\\
		\midrule[1pt]
		\multirow{3}*{UbuntuV2} & train & 1.42 & 3.24 & 0.29 & 68.92 &22.19& 6.65& 2.24\\
		~ & valid& 1.39 & 3.09 & 0.25  &70.78& 20.90& 6.47& 1.85\\
		~ & test & 1.39 & 3.13 & 0.25 & 70.69& 20.89& 6.18& 2.23\\
		\hline
		\multirow{3}*{DSTC7}  &train &1.40 &4.45  &0.38  &67.75& 25.37& 5.48& 1.40\\
		~ & valid& 1.39 & 4.42 & 0.37 & 67.76& 25.74& 5.44& 1.06\\
		~ & test& 1.45 & 4.42 & 0.42 &65.00& 26.10& 7.70& 1.20\\
		\hline
		\multirow{3}*{DSTC8*} &train & 3.82 & 24.70 & 5.39  & 2.96& 2.59& 3.85& 90.61\\
		~ & valid& 3.80 & 24.46 & 5.37 & 3.32& 3.09& 3.64& 89.95\\
		~ & test& 3.81 & 24.53 & 5.34 & 3.09& 2.76& 4.06& 90.09\\
		\bottomrule[1pt]
	\end{tabular}
	\caption{Statistics on extraction results. avg\#thd refers to the average number of threads per dialogue, avg\#turn refers to the average number of turns in each thread, and std\#turn refers to the average standard deviation of the number of turns in each thread per dialogue. 1-thd to 4-thd refers to the percentage of the number of dialogues with 1 to 4 threads in corresponding datasets.}
	\label{tab:parstatis}
\end{table*}
We first evaluate the extraction results on UbuntuV2, DSTC7 and DSTC8* with three metrics:
The average number of threads (\textbf{avg\#thd}) is to show how many dialogue threads are discovered in each dialogue, which ranges from 1 to 4. We didn't take all of the extracted threads into consideration, serving as a hard cut for the trade-off between information loss and memory usage of the model.
The average number of turns in each thread (\textbf{avg\#turn}) and the average standard deviation of the number of turns in each thread (\textbf{std\#turn}) are to measure the length of each thread. Dialogues context is not well separated if the length of each thread varies a lot~(i.e., the std\#turn is too high).

We apply the dialogue extraction algorithm in Section \ref{sec:DSA} on the three datasets. The statistics of extracted threads are in Table \ref{tab:parstatis}.
Firstly, we can find that the average number of threads is around $3.81$ for DSTC8* dataset while around $1.40$ for the other two datasets, which well aligns with the empirical observation that two-party dialogues tend to have more concentrated discussions with a smaller number of threads while multi-party dialogues usually contain more threads to accommodate conversation with high diversity. Also, as is listed in Table \ref{tab:dataset}, the number of turns for DSTC8* dataset is usually larger than UbuntuV2 and DSTC7 dataset, which naturally leads to more leaf nodes hence a larger number of threads. 
Secondly, the average length of threads is around $24.50$ for DSTC8* dataset while around $4.0$ for DSTC7 dataset and UbuntuV2 and the standard deviation for DSTC8* dataset is also larger. It shows that when the number of dialogue threads increases, the standard deviation of the length of each thread also tends to increase since some dialogue threads may catch more attentions while others may be ignored. 
In summary, DSTC8* is a more challenging multi-party dialogue dataset for dialogue context modeling than two-party dialogue datasets, including UbuntuV2 and DSTC7.

\subsection{Response Selection Results}
The response selection results of our Thread-Encoder models, including Thread-bi and Thread-poly, are shown in Table \ref{tab:V2D7} for UbuntuV2 and DSTC7 datasets, and in Table \ref{tab:discussion} for DSTC8*. 
\begin{table*}[ht!]
	\centering
	\small
	\begin{tabular}{l|cccc|cccc}
		\toprule[1pt]
		\multirow{2}{*}{} &
		\multicolumn{4}{c|}{UbuntuV2} &
		\multicolumn{4}{c}{DSCT7}\\ 
		Model &  hits@1 & hits@2 & hits@5 & MRR& hits@1 & hits@10 & hits@50 & MRR\\
		\midrule[1pt]
		DAM & - & -& -& - &  34.7 & 66.3 & - & 35.6\\
		ESIM-18 &  73.4 & 85.4 & 96.7 & 83.1 & 50.1 & 78.3 & 95.4 & 59.3 \\
		ESIM-19& 73.4 & 86.6& 97.4& 83.5 & 64.5 & 90.2 & \bf 99.4 & 73.5 \\
		IMN &  77.1 & 88.6 & 97.9 &- & -& -& -&- \\
		Bi-Enc & 83.6 & - &  98.8 & 90.1  &70.9 & 90.6 &  - & 78.1\\
		Poly-Enc & 83.9 & - & 98.8 & 90.3 &70.9 & 91.5 & - & 78.0\\
		Cross-Enc & \bf 86.5 & - & \bf 99.1 & \bf 91.9&71.7 & 92.4 & - & 79.0 \\
		\hline
		Thread-bi	& 83.8 &  92.4 & 98.5 & 90.0 & \bf 73.3$^\star$ &92.5 & 99.3 & 80.2$^\star$ \\
		Thread-poly & 83.6&\bf 92.5 &98.5 &90.0 &73.2$^\star$& \bf  93.6$^\star$&99.1& \bf80.4$^\star$\\
		\bottomrule[1pt]
	\end{tabular}
	\caption{Results on UbuntuV2 and DSTC7 dataset. Scores marked with $^\star$ are statistically significantly better than the state-of-the-art 
		with $p<0.05$ according to t-test.}
	\label{tab:V2D7}
\end{table*}

Since UbuntuV2 is too large, we only fine-tuned on this dataset for three epochs due to limited computing resources.
The performance of our model is similar to Bi-Enc and Poly-Enc on UbuntuV2.
Although the Cross-Enc rank top on UbuntuV2, it is too time-consuming and 
not practical~\cite{humeau2019poly}. It runs over 150 times slower than both Bi-Enc 
and Poly-Enc. Our model, Thread-bi, takes the top four threads~(see \secref{sec:number} for more details) 
into consideration with the inference time overhead similar to Bi-Enc and Poly-Enc. 
Besides, the reason why our model seems slightly worse than Poly-Enc is that UbuntuV2 is an easier dataset with fewer turns and threads according to Table \ref{tab:dataset} and Table \ref{tab:parstatis}. Consequently, our model degenerates towards Bi-Enc and Poly-Enc, and all four models~(Bi-Enc, Poly-Enc, Thread-bi, Thread-poly) actually yield similar results, with p-value greater than 0.05.

Due to the huge advancement of pre-trained models over other models shown on UbuntuV2 and DSTC7, we mainly compared the competitive state-of-the-art pre-trained models on DSTC8* dataset for through comparison as shown in Table \ref{tab:discussion}.
Our models achieve the new state-of-the-art results on both DSTC7 and DSTC8* dataset proving that threads based on dependency relation between turns are helpful for dialogue context modeling. 
We can see that using multiple vectors works much better than using only one representation. The gap between these two aggregation methods is not clear on UbuntuV2 and DSTC7, but much more significant on DSTC8* where the dialogues between multiple participants are much more complicated. This finding hasn't been shown in Humeau's work~\shortcite{humeau2019poly}. Besides, our model can enhance both kinds of pre-trained dialogue models on the multi-turn response selection task by comparing Thread-bi with Bi-enc and Thread-poly with Poly-enc.

It should be noted that the inherent properties of these three datasets are different according to \secref{sec:extrresults}. UbuntuV2 and DSTC7 datasets are dialogues between two parties, while DSTC8* dataset involves more complicated multi-party dialogue. This reveals that Thread-Encoder not only works under simple scenarios such as private chats between friends, but also acquires further enhancement under more interlaced scenarios such as chaos chat rooms.

\begin{table}[th!]
	\centering
	\small
	\begin{tabular}{lccc}
		\toprule[1pt]
		Model & \#Para & Train(h) & Test(\#dialog/s)   \\
		\midrule[1pt]
		Bi-Enc & 256.08M& 10.22 & 6.79  \\		
		Poly-Enc & 256.13M & 12.34 & 4.78 \\
		\hline
		Thread-bi &  256.08M& 16.36 &  4.73  \\
		Thread-poly &256.13M & 17.09 & 4.77  \\
		\bottomrule[1pt]
	\end{tabular}
	\caption{Total number of parameters, training time (h) and testing speed(\#dialogues per second) on DSTC8* main models.}
	\label{tab:dstc8}
\end{table}

The number of parameters, training time and testing speed are shown in Table \ref{tab:dstc8}. It takes more epochs for our model to convergence, while the testing speed is similar to Poly-Enc.

\subsection{Discussions on Model Design}

To further understand the design of our full model, we did several ablations on DSTC8*. All of the ablation results as listed in Table \ref{tab:discussion}. The descriptions and analysis are in following subsections.
\begin{table*}[th]
	\centering
	\small
	\begin{tabular}{lccccccccc}
		\toprule[1pt]
		ID &Method & Aggregation Type & Thread Type& \#Thread & hits@1 & hits@5 &hits@10 & hits@50 & MRR \\
		\midrule[1pt]
		1&\underline{Bi-Enc} & Average& Full-hty & 1& \underline{22.2} & \underline{43.0} & \underline{54.2} & \underline{88.7} & \underline{32.9} \\
		2&\underline{Poly-Enc} & Attention & Full-hty & 1 & \underline{32.5} & \underline{54.1} & \underline{64.4} & \underline{91.4} & \underline{43.1}  \\
		\midrule[0.5pt]
		3&Thread-bi& Average & Dep-extr & 1 & 20.2& 39.6& 51.1& 86.1& 30.5\\
		4&Thread-bi& Average & Dep-extr & 2 & 22.6& 42.5& 53.1&87.9 &32.9 \\
		5&\underline{Thread-bi}& Average & Dep-extr & 3 & \underline{23.4}& \underline{43.0}& \underline{54.9}& \underline{88.2}& \underline{33.8}\\
		6&Thread-bi& Average & Dep-extr & 4 &  22.9& 43.3& 55.1& 88.5& 33.5\\
		7&Thread-bi& Average & Dist-seg & 3 & 21.7 &43.2 & 55.2&88.8 &32.8 \\
		\midrule[0.5pt]
		8&Thread-poly& Attention & Dep-extr & 1 &29.4 &49.1 & 59.4& 88.7&39.5 \\
		9&Thread-poly& Attention & Dep-extr & 2 &32.0 &53.2 &63.2 &91.1 &42.5 \\
		10&Thread-poly& Attention & Dep-extr & 3 & 33.1 & 54.1& 64.2& 92.0 & 43.5 \\
		11&\underline{Thread-poly}& Attention & Dep-extr & 4 &  \underline{\bf 33.5$^\star$}& \underline{\bf 54.5$^\star$}& \underline{\bf 64.5}& \underline{91.7}& \underline{\bf 44.0$^\star$}\\
		12&Thread-poly& Attention & Dist-seg & 4 &33.2 &53.5 & 63.6&\bf 92.3$^\star$& 43.4\\
		\bottomrule[1pt]
	\end{tabular}
	\caption{Main results of DSTC8* (underlined) and ablation tests on DSTC8*. Scores marked with $^\star$ are statistically significantly better than Poly-Enc 
		with $p<0.05$ according to t-test.}
	\label{tab:discussion}
\end{table*}

\subsubsection{Different ways to generate threads}
\label{sec:ways}
We evaluate some reasonable alternative methods to extract dialogue threads
from the history, i.e.``Thread Type'' in Table \ref{tab:discussion}. 
\begin{itemize}
	\item \textbf{Full-hty} concatenate the full dialogue history in one thread. 
Our model degrades to Bi-Enc and Poly-Enc.
	\item \textbf{Dist-seg} segments the turns based on their distance to the next response. This idea is based on the intuition that the adjacent turns are possible to have strong connections. For example, if we use 4 threads, the dialogue in Figure \ref{fig:algorithm} will be segmented into $\langle\langle t_6, t_7\rangle, \langle t_4, t_5\rangle, \langle t_2, t_3\rangle, \langle t_1\rangle\rangle$.
	\item \textbf{Dep-extr} refers to the threads extraction procedure as explained in \algoref{alg:DSA}. 
\end{itemize}

Comparing in group ID-$\{1, 5, 7\}$ and ID-$\{2, 11, 12\}$, we get the following observations:
(1) Our extraction operations help with the response selection as both ID-$5$ and ID-$11$ have significant improvement despite the distance-based extraction method is a strong baseline. The dependency relations capture salient information in dialogue more accurately and yields better performance.
(2) Segmenting dialogues simply based on distance may hurt the storyline for each sub dialogue as ID-$7$ is worse than ID-$\{1, 5\}$, which hurts the representation ability of language models. 
(3) The information loss caused by Dist-seg can be partially made up by ``poly'' settings as ID-$12$ lies between ID-$2$ and ID-$11$. Generating multiple representations by aggregators may help to get multiple focuses in each thread. Thus interleaved sub-dialogues can be captured more or less. The gap between Dist-seg and Dep-extr will definitely be widened by improving the performance of sub-dialogue extraction.

\subsubsection{The number of threads to use}
\label{sec:number}

After deciding the way for extraction, the number of threads (i.e., \#Thread in Table \ref{tab:discussion}) to use is another key hyper-parameter for this model design. 

We tested our model using the number of threads ranging from $1$ to $4$. The results are shown in ID-$\{3\sim6\}$ and ID-$\{8\sim11\}$ from Table \ref{tab:discussion},  we draw following conclusions.
First, by comparing the results with only $1$ thread, we can see ID-$3$ and ID-$8$ are worse than Bi-enc and Poly-enc respectively. It shows that there does exist many cases that correct candidates that do not respond to the nearest dialogue threads. Considering only the nearest sub-dialogue is not enough.
Second, with the increasing number of threads from $1$ to $4$, the results go up and down for Thread-bi. The peak value is achieved when \#Thread equals $3$. Although more than $90\%$ of dialogues can be extracted into $4$ threads according to Table \ref{tab:parstatis}, the results doesn't go up with one more thread.
Some redundant dialogue threads far from the current utterances may bring noises for response selection.
Also, the negative effects of redundant dialogue threads for Thread-poly reflect on the limited improvements and even decreases on hits@50 between ID-$10$ and ID-$11$. 
Designing a metric to filter the extracted dialogue threads automatically is our future work.

%% file: relatedwork.tex
\section{Related Work}
\label{sec:relatedwork}
Related work contains dialogue dependency parsing and multi-turn response selection.

\subsection{Dialogue dependency parsing}

Discourse parsing has been researched by scientists especially in linguistics for decades. 
Asher and Lascarides~\shortcite{0031949} proposed the SDRT theory with the STAC Corpus~\cite{AsherHMBA16} which made a great contribution to the discourse parsing on multi-party dialogues. Shi and Huang~\shortcite{ShiH19} proposed a sequential neural network and achieved the state-of-the-art results on this dataset. Another similar task is dialogue disentanglement~\cite{DuPX17}. This task isn't focusing on developing discourse theories but trying to segment the long dialogues according to topics. It takes each turn in the dialogue as a unit, and only care about whether there is a relation between two turns, which is called ``reply-to'' relation. Due to the scarcity of annotated dialogues across domains under SDRT theory, the predicted dependency relations had never been used for down-streaming tasks, such as response selection and dialogue summarization. In this paper, we take advantage of both the simplicity of the ``reply-to'' relation and the sequential parsing methods~\cite{ShiH19} to do dialogue dependency parsing. Developing general discourse parsing with relations types and take relation types into consideration may be future work.

\subsection{Multi-turn response selection}
\label{sec:mtrs}
Multi-turn response selection task was proposed by Lowe et al.~\shortcite{LowePSP15} and the solutions for this task can be classified into two categories: the sequential models and the hierarchical models. To begin with, the sequential models~\cite{LowePSP15} were directly copied from the single-turn response selection task since we can regard the multiple history turns as a long single turn.  Considering the multi-turn characteristic, Wu et al.~\shortcite{WuWXZL17} proposed the sequential matching network (SMN), a new architecture to capture the relationship among turns and important contextual information. SMN beats the previous sequential models and raises a popularity of such hierarchical models, including DUA~\cite{ZhangLZZL18}, DAM~\cite{WuLCZDYZL18}, IOI~\cite{TaoWXHZY19}, etc. The ESIM~\cite{abs-1802-02614}, which is mainly based on the self and cross attention mechanisms and incorporates different kinds of pre-trained word embedding. It changed the inferior position of the sequential model, making it hard to say which kind of architecture is better.

Due to the popularity of the pre-trained language models such as BERT~\cite{DevlinCLT19} and GPT~\cite{radford2018improving}, the state-of-the-art performance on this task was refreshed~\cite{vig2019comparison}. Work such as \cite{abs-1908-04812} and \cite{humeau2019poly} further shows that the response selection performance can be enhanced by further pretraining the language models on open domain dialogues such as Reddit~\cite{abs-1904-06472}, instead of single text corpus such as BooksCorpus~\cite{ZhuKZSUTF15}. These models can be also regarded as the sequential models because they concatenate all the history turns as the input to the model while ignoring the dependency relations among the turns. Inspired by these works, we incorporate the dependency information in the dialogue history into the response selection model with the pre-trained language model on dialogue dataset.

In this work, we focus on the effectiveness of exploiting dependency information for dialogue context modeling and follow the data preprocessing steps in two-party dialogue datasets, including UbuntuV2 and DSTC7, which have no special designs for speaker IDs. In the papers for DSTC8 response selection track, such as \cite{abs-2004-01940}, many heuristic rules based on speaker IDs are used for data preprocessing, which greatly helps to filter out unrelated utterances. However, they also definitely lead to losing some useful utterances. These hard rules will hurt the completeness of the meaning in each thread and are not suitable for us. As a result, the results on the response selection task for DSTC8 dataset are not comparable.
We will take advantage of the speaker information into both extraction and dialogue understanding models as our future work.

%% file: conclusion.tex
\section{Conclusion}
As far as we know, we are the first work bringing the dependency information of 
dialogues into the multi-turn response selection task. We proposed the dialogue extraction algorithm 
and Thread-Encoder model, which becomes the state-of-the-art on several well-known 
ubuntu datasets. In the future, we will move on to develop a more general 
dialogue dependency parser and better incorporate dependency information into 
dialogue context modeling tasks.